\documentclass[journal,comsoc]{IEEEtran}
\usepackage[T1]{fontenc}
\usepackage{cite}

\ifCLASSINFOpdf
   \usepackage[pdftex]{graphicx}
\else
   \usepackage[dvips]{graphicx}
\fi

\usepackage{amsmath}
\usepackage[cmintegrals]{newtxmath}

\usepackage[numbers]{natbib}
\usepackage{booktabs}
\usepackage{multirow}
\usepackage{multicol}
\usepackage{amsmath}
\usepackage{algpseudocode}
\usepackage{mathtools}
\usepackage[linesnumbered,ruled,vlined]{algorithm2e}
\usepackage{hyperref}

\mathchardef\mhyphen="2D

\begin{document}
%
\title{Method-Based Reasoning for Large Language Models: Extraction, Reuse, and Continuous Improvement}

\author{Hong~Su
\IEEEcompsocitemizethanks{\IEEEcompsocthanksitem H. Su is with the School of Computer Science, Chengdu University of Information Technology, Chengdu, China.\\
 E-mail: suguest@126.com. \\
\protect\\
}
\thanks{}}

\markboth{Journal of \LaTeX\ Class Files,~Vol.~14, No.~8, August~2015}%
{Shell \MakeLowercase{\textit{et al.}}: Bare Demo of IEEEtran.cls for IEEE Communications Society Journals}
%

\maketitle

\begin{abstract}
Large language models (LLMs) have shown impressive capabilities across a wide range of language tasks. However, their reasoning process is primarily guided by statistical patterns in training data, which limits their ability to handle novel problems and perform consistent logical reasoning. In this paper, we propose a method-based model that enhances LLMs with explicit, reusable procedures extracted from training content, generated responses, and user interactions. Each method is represented as a pair consisting of a problem and its corresponding solution, stored externally and ranked based on feedback. When a new query is received, the system retrieves and applies the most relevant methods to guide the LLM’s response. Our model enables continual learning, method reuse, and logical consistency beyond next-token prediction. Experimental results demonstrate that the system improves factual verification and generalization in complex prompts, and that newly learned methods can outperform earlier ones through user-driven refinement. 
\end{abstract}


\IEEEpeerreviewmaketitle

\section{Introduction}

Large language models (LLMs) have achieved remarkable success in a wide range of natural language processing tasks, including text generation \cite{kumichev2024medsyn}, question answering \cite{zhuang2023toolqa}, and dialogue systems \cite{chen2024structured}. These models are typically built on Transformer architectures \cite{mo2024large} \cite{liu2025attention}, which learn statistical relationships between words by training on massive corpora of text data. During inference, the model predicts the most likely next token given a sequence of known tokens, leveraging patterns learned from its training data \cite{shen2025efficient}.

Despite their success, LLMs face two fundamental limitations. First, their behavior is primarily driven by common statistical patterns in language, which makes it difficult to handle novel problems that deviate from previously seen distributions. When the input falls outside the scope of pre-trained patterns, the model may struggle to produce correct or coherent outputs. Second, the generation process in LLMs lacks explicit logical reasoning \cite{cheng2025empowering}. Although logical structure may sometimes emerge implicitly through statistical learning, LLMs often fail to verify the validity of their outputs from a reasoning standpoint. As a result, logical errors—such as hallucinating unsupported facts or making inconsistent assumptions—can go undetected.

To address these limitations, we propose a method-oriented approach that introduces an explicit layer of procedural reasoning on top of LLMs. In this model, a method is defined as a pair consisting of a problem and its corresponding solution procedure. By extracting and applying such methods, the system can reason more consistently and adapt known procedures to solve structurally similar but previously unseen problems. This mirrors how humans apply learned strategies to new situations—for instance, a student applying a familiar technique to solve a new type of math problem.

While existing techniques such as Chain-of-Thought (CoT) prompting~\cite{wei2022chain} aim to improve LLM reasoning by encouraging step-by-step thinking, they rely on manually designed prompt templates and are not content-adaptive. CoT is effective in some domains but not universally applicable across all types of user inputs. In contrast, our method-based approach learns directly from LLM training content, outputs, and user interactions, enabling dynamic and context-sensitive method acquisition.

In this paper, we propose a model for automatic method extraction and reuse within LLM systems. Methods are automatically extracted from training materials, generated responses, or user inputs. Each method is represented as a problem-solution pair and stored in a method management module. When a new user query is received, the system compares it to stored problems to retrieve candidate methods. These candidates are passed to the LLM for execution, and the resulting outputs are ranked by the user or system. Over time, higher-ranked methods are favored, enabling continual improvement of reasoning quality.

The key contributions of this work are summarized as follows. First, we introduce a method-based reasoning model that enhances LLM output by extracting reusable procedures from interactions. Second, we propose a structure for representing and storing methods as problem-solution pairs, enabling semantic matching and generalization. Third, we develop an automatic method extraction and ranking mechanism that requires no user intervention and adapts dynamically over time. Lastly, we demonstrate through experimental evaluation that our approach improves the logical reliability and adaptability of LLM-generated outputs.

The remainder of this paper is organized as follows. Section~\ref{sec_related_work} reviews related work. Section~\ref{sec_think_model} introduces the proposed method-based reasoning model. Section~\ref{sec_verification} presents our experimental verification and analysis. Finally, Section~\ref{sec_conclusion} concludes the paper with a summary and future directions.

\section{Related Work} \label{sec_related_work}

Recent research has increasingly explored the enhancement of large language models (LLMs) through structured reasoning, prompt engineering, and interaction-driven adaptation. Our work contributes to this growing body of research by introducing a method-based model that emphasizes reusable problem-solving logic extracted from LLM interactions.

\textbf{Prompt Engineering and Chain-of-Thought Reasoning.}  
Prompt-based methods such as Chain-of-Thought (CoT)~\cite{wei2022chain} and ReAct~\cite{yao2023react} have demonstrated that explicitly guiding LLMs through intermediate reasoning steps improves accuracy in complex question answering, mathematical problem-solving, and planning tasks. CoT operates by prompting the model to “think aloud” before providing a final answer, which often improves reasoning transparency and interpretability. Similarly, ReAct combines reasoning with action, encouraging LLMs to interleave environment interaction (e.g., tool use) with internal reasoning.

Despite their effectiveness, both CoT and ReAct rely heavily on carefully crafted prompt templates that are tailored to specific domains or task types. This limits scalability and reuse, particularly when the domain or logic evolves over time. Our work complements and extends these approaches by introducing a mechanism that automatically extracts generalized reasoning patterns (methods) from LLM outputs. These methods are stored structurally and reused across sessions, reducing dependence on human-crafted prompts and supporting continual improvement through interaction.

Moreover, while CoT typically focuses on decomposing a single task into steps, our model extracts reusable methods that capture the entire solution strategy for a class of problems. This allows for method transfer and modular composition, enabling the system to build a repertoire of reusable, abstract procedures.

\vspace{0.5em}
\textbf{Reinforcement Learning from Human Feedback.}  
Reinforcement Learning from Human Feedback (RLHF)~\cite{ouyang2022training} has become a standard mechanism for aligning LLM outputs with human preferences. In systems such as InstructGPT and ChatGPT, RLHF is used to reward helpful, harmless, and honest responses through ranked comparisons and preference modeling. However, the focus is typically on tuning the language model’s output distribution rather than structuring or storing the reasoning behind preferred outputs.

Our model integrates RLHF-like feedback mechanisms not just to adjust output selection, but to refine and update a library of reusable methods. When users evaluate multiple candidate solutions, their rankings are incorporated as external scores that inform method selection, replacement, and retention. This approach blends symbolic reasoning with preference modeling, enabling the system to prioritize logic patterns that are repeatedly validated by human users.

In contrast to model-level alignment, our method-based feedback loop operates at the reasoning structure level, enabling explainable and auditable improvements over time. This is especially valuable in domains where logic, rules, or verification steps must be made explicit, such as legal analysis, software configuration, or scientific inquiry.

\vspace{0.5em}
\textbf{Retrieval-Augmented Generation (RAG).}  
Retrieval-Augmented Generation (RAG)~\cite{lewis2020retrieval} \cite{zhao2024retrieval} augments LLMs with access to an external document store or knowledge base. At inference time, relevant documents are retrieved and incorporated into the prompt context to help the LLM generate grounded, factually accurate answers \cite{fan2024survey}. This hybrid approach improves factuality and allows the model to operate with smaller parameter sizes while maintaining strong performance.

While RAG systems retrieve and cite external factual content, our model retrieves abstract methods—structured problem-solving procedures extracted from previous interactions. These methods are not mere references, but logic-aware programs or stepwise strategies that can be applied dynamically to new problems. This distinction is critical: instead of enhancing the factual grounding of a single output, we enhance the logical reliability and reusability of entire solution paths.

Additionally, unlike typical RAG pipelines that use dense vector similarity to retrieve documents, our system incorporates both feature-based similarity and user feedback ranking to retrieve and apply methods. This adds an extra layer of interpretability and control, making it suitable for domains that require repeatable and verifiable decision-making processes.

\textbf{External Memory and Tool-Augmented LLMs.}  
Recent work has explored enhancing LLMs by equipping them with the ability to interact with external tools or memory systems. Toolformer~\cite{schick2023toolformer}, for example, fine-tunes models to decide when and how to call APIs, allowing the LLM to autonomously use external tools such as calculators, search engines, or weather services. Similarly, models like LlamaIndex \cite{zirnstein2023extended} and LangChain \cite{topsakal2023creating} provide structured access to external databases or memory indices, allowing LLMs to retrieve long-term context or document-level knowledge.

These systems focus on augmenting LLMs with access to external factual or functional resources. In contrast, our method management module acts as an \textit{external procedural memory}—a repository of reasoning strategies rather than data or facts. Instead of remembering specific facts, it remembers \textit{how to solve problems}, which includes preconditions, steps, and conditional actions.

This procedural memory enables logic-level consistency across sessions and tasks. It supports ranking, replacing, and reusing methods based on user feedback, thereby offering a bridge between symbolic planning (e.g., classical AI-style procedural knowledge) and neural language generation. While existing tool-augmented systems aim to fill content gaps, our model aims to close reasoning gaps by supplying verified procedural knowledge learned from prior LLM interactions.

\vspace{0.5em}
\textbf{Self-Improving and Continual Learning Systems.}  
Several studies have proposed architectures for self-improving LLMs, where the model iteratively learns from its own outputs \cite{ji2025unlocking}. Notable examples include models for self-refinement~\cite{huang2022self}, bootstrapping, or on-the-fly code repair, where outputs are recycled as new training data or refined via meta-prompting.

However, most of these approaches treat improvement as an implicit fine-tuning or output filtering step, rather than as structured method acquisition. They typically lack a persistent, interpretable memory of the procedures behind those improvements, making it difficult to reuse or transfer logic to new domains.

Our model complements these efforts by formalizing the notion of method as a reusable unit of reasoning. Methods are extracted from LLM interactions, stored explicitly, associated with problem features, and refined through user ranking. This introduces an interpretable and auditable mechanism for continual logical improvement—analogous to building a method library that grows with each session.

Moreover, unlike pure self-training systems which may drift without human oversight, our architecture remains grounded through user feedback (via ranking) and method filtering (via structure checks). This hybrid approach balances autonomy with control, allowing both safe refinement and transferability to unseen problems with similar logical form.

In summary, our work is positioned at the intersection of retrieval-based reasoning, logic-aware prompting, and memory-augmented interaction. It contributes a novel mechanism for dynamically growing and refining a repository of executable methods that improves LLM performance across time and users.

\section{Method-Based Reasoning for Large Language Models} \label{sec_think_model}


In conventional LLM usage, training and inference are primarily regarded as processes for generating conversational or textual content. In contrast, this work reinterprets the training material as a repository of \textit{methods}—structured procedures that can be extracted and reused to solve new tasks—rather than merely viewing it as a collection of words, sentences, or paragraphs.

A method refers to a procedure composed of one or more actions aimed at solving a specific problem. Unlike static text generation, method extraction operates at a semantic level, focusing on the underlying logic rather than on surface-level word predictions. This abstraction allows extracted methods to be transferred and adapted to similar problems, even when the textual representations differ.

Method selection and optimization are guided by logical reasoning or preference mechanisms, such as reinforcement learning from human feedback (RLHF), rather than by conventional token-level prediction. Further details on method ranking are discussed in Section ~\ref{sec_meas_meth}.

\begin{figure}
    \includegraphics[width=3.5in]{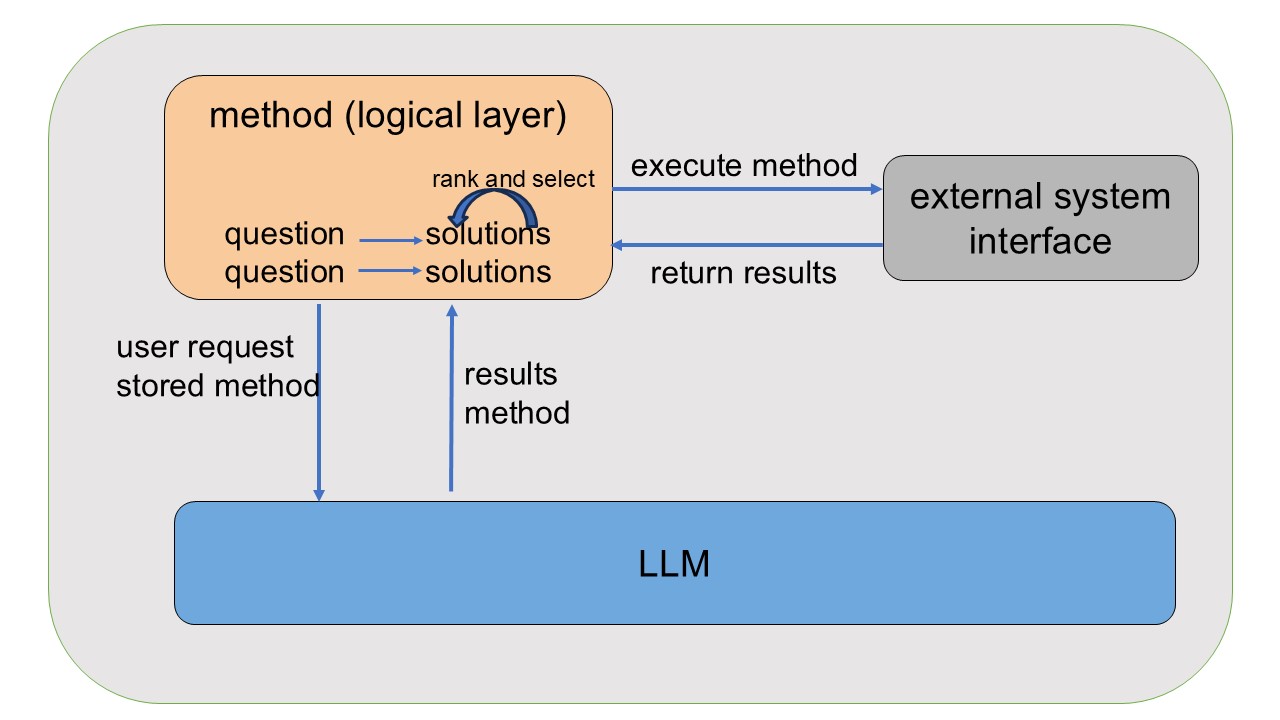}
    \caption{The high-level diagram of the method-based LLM model}
    \label{architecture}
\end{figure}

In the proposed LLM-based method extraction model, the LLM is utilized to identify and extract actionable procedures from input content, as illustrated in Figure~\ref{architecture}. These extracted methods may take the form of textual descriptions—serving as prompts or hints to guide subsequent LLM behavior—or executable external processes such as Python scripts or API calls, as detailed in Section~\ref{sec_method_format}.

A method is characterized by a logical pairing of a \textit{problem} and its corresponding \textit{solution}. The problem describes the objective or issue to be resolved, while the solution consists of one or more actions taken to address it. This is formally represented in Equation~\eqref{eq_method_definition}, where a method is defined as a tuple of logical components:

\begin{equation} \label{eq_method_definition}
\text{Method} = \{ \textit{problem}, \textit{solution} \}
\end{equation}
 , where $\textit{problem}$ and $\textit{solution}$ denote the abstract representations of the problem and solution, respectively. For brevity, they may be referred to simply as \textit{p} and \textit{s} when no ambiguity arises.

This method-centric approach differs fundamentally from storing all content or internal relationships (e.g., query-key-value matrices in Transformer models). Simply preserving all training materials may result in a bloated memory space, making it difficult for the LLM to discern which parts are relevant to a new problem. In contrast, explicitly extracting and storing logical methods provides a more structured and reusable mechanism for generalization.

To determine whether a piece of material constitutes a method, the LLM is prompted to assess its suitability. If deemed a method, the LLM is then required to identify both the underlying problem and the corresponding solution. The problem can be derived by asking the LLM to formally describe the task, summarize the issue, or classify the type of question. In some cases, the original text from which the method was extracted may also be used directly as the problem description.

The problem component can take various forms—not limited to a single word—but potentially spanning a sentence, a paragraph, or multiple paragraphs. Importantly, the problem need not be an exact identifier (e.g., an exact name); rather, it may represent a semantically vague or generalized formulation. This enables the LLM to recognize and associate similar but non-identical problems through high-level feature alignment.

By decoupling problems and solutions, this model enables the construction of a flexible mapping between abstract problem representations and corresponding solutions. Formally, we define a many-to-many mapping function $\mathcal{M}$:

\begin{equation} \label{eq:problem_solution_mapping}
\mathcal{M}: \mathcal{P} \rightarrow 2^{\mathcal{S}}, \quad \mathcal{M}(p_i) = \{ s_j \mid s_j \text{ solves } p_i \}
\end{equation}
 , where $\mathcal{P}$ is the set of all problems and $\mathcal{S}$ is the set of all solutions. Each problem $p_i \in \mathcal{P}$ may have multiple associated solutions $s_j \in \mathcal{S}$, and vice versa. This many-to-many relationship forms the basis for method composition and reuse.

To support scalable retrieval, we organize problems hierarchically using a tree structure $\mathcal{T} = (V, E)$, where each node $v \in V$ represents a problem (or a problem cluster), and each edge $(v_i, v_j) \in E$ represents a semantic relationship (e.g., generalization or specialization):

\begin{equation} \label{eq:tree_structure}
\mathcal{T}: \text{Root} \rightarrow \text{Children}(p) \rightarrow \text{Leaves}, \quad \text{with } p \in \mathcal{P}
\end{equation}

Each node $v$ stores a tuple $(p, \mathcal{M}(p))$, linking a problem to its set of solutions. Traversal of this tree allows for efficient discovery of relevant methods even in large and diverse knowledge bases.

This hierarchical representation supports logical reasoning, method reuse, and efficient similarity-based retrieval, especially when combined with LLM-generated embeddings or semantic clustering.

\subsection{Storage of Methods}
\subsubsection{Method Storage Tree}
The tree used to store methods are called \textbf{method storage tree}, in which methods are organized according to their associated problems.

The tree structure is built primarily based on the problem descriptions, allowing similar problems to be grouped together. Although solutions may also be used for organization, they are discussed separately under section of \ref{sec_method_storage} and \ref{sec_solution_part}. The process begins with an initially empty tree. When a new method is introduced, the LLM is prompted to determine the appropriate location in the existing tree for insertion. Or if there are some methods, the LLM is then used to group existing methods into a tree based on problem similarity.

Each node in the tree consists of a problem and its associated solutions. Since a problem may have multiple valid solutions, each node can store a list of method variants, as illustrated in Equation~\eqref{eq_question_method}.

\begin{equation} \label{eq_question_method}
\textit{question} \rightarrow \{ \textit{s}_a, \textit{s}_b, \cdots \}
\end{equation}

If the root node of the tree contains a method, it may serve as a general-purpose method applicable across a wide range of scenarios. In cases where method ranking is unavailable, all methods along the path from the root to a specific leaf node can be provided to the LLM, which then selects the most suitable one.

Another important consideration is the \textbf{scope} of method storage. Storage may be maintained either at the \textit{user level} or the \textit{LLM level}. User-level storage retains methods and associated problems for individual users, allowing personalized learning and reuse. In contrast, LLM-level storage aggregates methods extracted from all users, including those derived from training data and user interactions. New methods, once extracted, can be ranked later to determine their utility.

\subsubsection{Methods for Improving Other Methods} \label{sec_method_storage}
In our model, a method (or solution) is typically associated with solving a specific user problem. However, there are scenarios where a method itself may become the subject of another method—especially when the original method fails, produces suboptimal results, or requires refinement.

For instance, if a method produces an incorrect or incomplete solution, another method can be invoked to validate, revise, or enhance it. One common example is the use of Chain-of-Thought (CoT) prompting to transform a shallow method into a more structured, step-by-step reasoning process. This type of meta-method can be seen as a higher-level strategy that operates on other methods rather than on raw input problems.

We formalize this idea as follows. Let $\mathcal{M}$ denote the set of all methods, and let $\text{Apply}(m, q)$ denote the application of method $m$ to a query $q$ or a method $m$ to produce an output. Then, for a method $m_1 \in \mathcal{M}$ and an input $q$, we define its output as:

\begin{equation}
    o_1 = \text{Apply}(m_1, q)
\end{equation}

If the output $o_1$ is deemed unreliable, a secondary method $m_2$ can be applied to the original method $m_1$ (with knowledge of $q$) to improve its behavior:

\begin{equation} \label{eq_m4f}
    o_2 = \text{Apply}(m_2, m_1, q)
\end{equation}

Here, $m_2$ is acting on $m_1$ as its input context—serving as a higher-level reasoning strategy. It may, for instance, prompt the system to re-express $m_1$ using CoT or add a validation step before accepting $m_1$'s result.

Thus, a hierarchical structure emerges, where:

- Leaf nodes represent base-level methods that address user-level queries.

- Internal nodes represent higher-level methods that operate on other methods, refining or validating their behavior.

We define a meta-method relationship as follows:

\begin{equation}
    m_2 \Rightarrow m_1 \quad \text{(i.e., $m_2$ refines or validates $m_1$)}
\end{equation}

Thus, a hierarchical structure enables the system to evolve complex strategies by composing and improving methods over time. The architecture supports recursive reasoning, layered validation, and the ability to adapt existing procedures for improved performance and reliability. 

\subsubsection{Solution Division and Part-Wise Storage} \label{sec_solution_part}

In addition to storing complete solutions, we enable fine-grained storage and manipulation of individual solution components, referred to as \textit{solution parts}. A solution typically consists of multiple procedural steps, each corresponding to a distinct subtask. By decomposing solutions into parts, we support modular reuse and targeted refinement.

Formally, a solution is represented as an ordered sequence of parts:

\begin{equation} \label{eq_solution_sequence}
\text{solution} = \langle \text{solution}_{p1}, \text{solution}_{p2}, \dots, \text{solution}_{pn} \rangle
\end{equation}

Each part $\text{solution}_{pi}$ is associated with a specific functional role within the overall solution. The entire solution is indexed by its logical characteristics, while each part can be stored, retrieved, and updated independently.

To support refinement, we define a ranking function $\text{Rank}(\cdot)$ that scores candidate solution parts based on their quality or effectiveness. For a given position $i$, let $\mathcal{C}_i$ denote the set of candidate parts. The updated part at position $i$ is selected by:

\begin{equation} \label{eq_solution_update}
\text{solution}_{pi}^{\text{new}} = \arg\max_{\text{candidate} \in \mathcal{C}_i} \text{Rank}(\text{candidate})
\end{equation}

The updated solution sequence then becomes:

\begin{equation} \label{eq_solution_updated}
\text{solution}^{\text{new}} = \langle \text{solution}_{p1}, \dots, \text{solution}_{pi}^{\text{new}}, \dots, \text{solution}_{pn} \rangle
\end{equation}

This part-wise update mechanism allows incremental improvements to solutions by replacing only underperforming components while retaining effective ones. It also facilitates generalization, as common solution parts can be reused across different tasks when subtasks overlap.

\subsection{Always-On Extraction: Automatic Method Mining}
To support continuous and automated method extraction, we adopt an \textit{always-on extraction} strategy. This approach attempts to identify methods from all available content sources, including:
(1) the LLM's training or fine-tuning data,
(2) the real-time outputs generated by the LLM, and
(3) external inputs such as user prompts or third-party API queries.

This strategy enables the system to capture new methods that may not be explicitly present in the original training set. For example, users may propose corrections or novel procedures during interaction with the LLM, and such content can be automatically mined to extract reusable methods. These user-derived methods are typically stored at the user level to personalize future interactions.

Automatic extraction can be implemented either within the LLM or via an external software module. In this work, we adopt the latter: a third-party software component continuously monitors input-output interactions and extracts candidate methods. This differs from preconfigured reasoning strategies such as Chain-of-Thought (CoT), which are static and predefined. In contrast, our method extraction system is dynamic and context-aware, adapting to diverse application scenarios.

Method extraction can be triggered either during model training (by analyzing mini-batches) or during user requirement time (by monitoring interactions). In the latter case, the system identifies methods in real time and stores them for future use. When a user interacts with the LLM, the third-party module retrieves potentially relevant stored methods and filters them using a ranking mechanism. Only high-confidence candidates are forwarded to the LLM. This two-stage process—external filtering followed by internal selection—is detailed further in Section~\ref{sec_meas_meth}.

Formally, let $\mathcal{D}_{\text{input}}$ denote the collection of all input materials, which may include training data, LLM outputs, and user prompts:

\begin{equation} \label{eq_input_data}
\mathcal{D}_{\text{input}} = \mathcal{D}_{\text{train}} \cup \mathcal{D}_{\text{output}} \cup \mathcal{D}_{\text{user}}
\end{equation}

The method extractor $\mathcal{E}$ is a function that scans $\mathcal{D}_{\text{input}}$ and produces a set of candidate methods:

\begin{equation} \label{eq_extractor}
\mathcal{M}_{\text{candidate}} = \mathcal{E}(\mathcal{D}_{\text{input}})
\end{equation}

Each extracted method $\text{method}_i \in \mathcal{M}_{\text{candidate}}$ is structured as:

\begin{equation} \label{eq_method_structure}
\text{method}_i = \{ \text{p}_i, \text{s}_i \}
\end{equation}

The candidates are then scored by a ranking function $\mathcal{R}$, which may reflect logical validity, relevance, or feedback:

\begin{equation} \label{eq_ranking}
\mathcal{M}_{\text{filtered}} = \{ \text{method}_i \in \mathcal{M}_{\text{candidate}} \mid \mathcal{R}(\text{method}_i) \geq \tau \}
\end{equation}

Here, $\tau$ denotes a tunable threshold that is used to filter out lower-quality methods based on their ranking scores. The resulting filtered set, denoted as $\mathcal{M}_{\text{filtered}}$, is then provided to the LLM for application to the user's query.

This filtering mechanism enables continuous refinement of the method set: newly discovered, higher-quality methods can progressively replace older, less effective ones. As a result, the system possesses the potential to evolve by learning and updating methods directly from ongoing user interactions and contextual content.

\subsection{Measurement of Methods} \label{sec_meas_meth}

Since multiple methods may be applicable to a given problem, it is necessary to define a mechanism for evaluating and selecting the most appropriate one. We adopt a dual-ranking strategy comprising two complementary components:

\begin{itemize}
    \item \textbf{External Ranking (User-Guided):} Based on reinforcement learning from human feedback (RLHF), users are presented with multiple candidate solutions and are asked to evaluate or rank them. Given that methods operate at a logical and semantic level, human judgment is particularly well-suited for assessing their quality and effectiveness.

    \item \textbf{Internal Ranking (LLM-Guided):} The LLM itself evaluates the candidate methods and selects the one it deems most appropriate. It also has the ability to output alternative methods or solutions, enabling the user to make a final selection if needed.
\end{itemize}

We formalize the ranking operation as follows. Given a set of candidate solutions:
\begin{equation}
\{ \text{s}_a, \text{s}_b, \cdots \}
\end{equation}

These are mapped to a ranked list:
\begin{equation}
\{ (\text{s}_a, \text{rank}_a), (\text{s}_b, \text{rank}_b), \cdots \}
\end{equation}

In practice, we employ a \textbf{two-step filtering and selection process} during user interaction with the LLM:

\begin{enumerate}
    \item \textbf{External Filtering:} The third-party software first applies external ranking to remove low-ranked or irrelevant methods. This stage retains a diverse subset of high-potential candidates while discarding clearly inferior ones. Notably, newly extracted methods (i.e., those without a rank) bypass this filter to allow exploration and learning.

    \item \textbf{Internal Selection:} The LLM then selects the most suitable method from the filtered candidates. It may also present additional alternatives for the user to compare, ensuring transparency and adaptability.
\end{enumerate}

This dual-ranking mechanism ensures both adaptability (via human feedback) and efficiency (via LLM inference), while maintaining a curated, high-quality method repository over time.

To formalize method selection at the logical level, we define a logical utility function $\mathcal{U}$ that evaluates a method $\text{method}_i$ with respect to a problem $\text{p}_j$:

\begin{equation} \label{eq_logical_utility}
\mathcal{U}(\text{method}_i, \text{p}_j) = \text{Relevance}(\text{p}_i, \text{p}_j) \cdot \text{Effectiveness}(\text{s}_i)
\end{equation}

Here, $\text{p}_i$ and $\text{s}_i$ are the components (question and solution) of method $i$, and:
\begin{itemize}
  \item $\text{Relevance}(\cdot, \cdot)$ measures the semantic or logical similarity between the problem addressed by the method and the current user problem.
  \item $\text{Effectiveness}(\cdot)$ estimates the success likelihood of the method's solution, potentially derived from prior usage or RLHF scores.
\end{itemize}

The optimal method is selected by maximizing this logical utility:

\begin{equation} \label{eq_best_method}
\text{method}^ = \arg\max_{\text{method}_i \in \mathcal{M}_{\text{filtered}}} \mathcal{U}(\text{method}_i, \text{p}_{\text{target}})
\end{equation}

In this model, the comparison and selection are made over the meanings and problem-solving intents of methods rather than surface-level lexical forms. This facilitates generalization, transfer, and reuse across tasks with structurally similar problems but varying textual expressions.

\subsection{Brief Proof: Why Methods Generalize to New Problems}

To justify the applicability of extracted methods to new problems, we consider both the structural nature of methods and their semantic decoupling from surface-level content.

A method is defined as a logical pair:
\begin{equation}
\text{method}_i = \{ \text{p}_i, \text{s}_i \}
\end{equation}
where $\text{s}_i$ represents a set of actions or procedures, and $\text{p}_i$ is a high-level abstraction of the issue being addressed.

While the solution component of a method often remains stable across similar problems, the problem component can vary significantly in phrasing or context. This decoupling enables a known method to be reused in scenarios where the surface expression of the problem differs, but the underlying logical structure remains consistent.

Formally, let $\text{p}_{\text{new}}$ be a new problem presented to the system. If there exists a method $\text{method}_i$ such that:

\begin{equation} \label{eq_generalization_condition}
\text{Relevance}(\text{p}_{\text{new}}, \text{p}_i) \geq \theta
\end{equation}

where $\theta$ is a predefined relevance threshold (e.g., based on semantic similarity or logical structure), then $\text{method}_i$ can be considered a viable candidate for solving $\text{p}_{\text{new}}$.

This generalization property mirrors human reasoning: once a method for solving a class of problems is learned (e.g., solving quadratic equations), it can be reused in varied situations with similar structure but different contextual phrasing.

Moreover, because methods are dynamically extracted from LLM-generated content or training materials, they inherently capture reusable structures from practical examples. For instance, strategies such as Chain-of-Thought (CoT) reasoning illustrate how a single method can generalize across a wide array of domains—ranging from mathematical proofs to planning tasks.

In summary, the core advantage of the method-based model lies in its abstraction:
\begin{itemize}
  \item Methods are not tied to specific content but to the logical structure of problem-solving.
  \item The problem component serves as a flexible matcher for identifying reuse opportunities.
  \item The solution component encodes reusable procedures that can be invoked across multiple scenarios.
\end{itemize}

This model allows for dynamic reuse, adaptation, and substitution—supporting a scalable approach to generalization beyond static training corpora.

\subsection{Procedure for Method Extraction and Application}
This section describes the end-to-end process of extracting and applying methods using large language models (LLMs), supported by an auxiliary component called the \textit{method extraction module}. This module facilitates the interaction between users and the LLM, maintains the method repository, and determines which method to apply in response to a given input.

When a user submits an input to the system, the method extraction module first prompts the LLM to analyze the input and extract its semantic or structural features. These features are then compared against those in the existing method storage tree, which is organized by problem descriptions. The system queries the LLM to assess whether any previously stored methods are applicable to the current input. If so, the corresponding methods are retrieved and ranked based on criteria such as relevance, similarity, or historical effectiveness.

If matching methods are found, the top-ranked candidates are passed back to the LLM. The LLM then generates distinct responses using each candidate method. These responses are optionally presented to the user, who may rank or select the best one. Once a suitable method has been applied successfully, the method extraction module stores the association between the input features and the applied solution, thus enriching the method repository.

In the event that no matching method is found, the system defaults to using generic or approximately similar methods. When multiple methods are equally applicable, the selection process may rely on configurable heuristics such as choosing the most recent method, randomly sampling, or generating and comparing multiple outputs for evaluation.

The entire procedure is summarized in Algorithm~\ref{alg:method_procedure}.

\begin{algorithm}[ht]
\caption{Method Extraction and Application Procedure}
\label{alg:method_procedure}
\KwIn{User input $x$, method repository $\mathcal{M}$}
\KwOut{Selected method and final output}

Extract feature representation $f_x \leftarrow \text{LLM.extract\_features}(x)$\;
Retrieve candidate methods $\mathcal{M}_c \leftarrow \text{match}(f_x, \mathcal{M})$\;
\If{$\mathcal{M}_c \neq \emptyset$}{
    Rank candidates: $\mathcal{M}_r \leftarrow \text{rank}(\mathcal{M}_c)$\;
    Send $\mathcal{M}_r$ to LLM to generate outputs\;
    User selects or ranks the best output\;
    Update repository: $\mathcal{M} \leftarrow \mathcal{M} \cup \{ (f_x, \text{best\_method}) \}$\;
}
\Else{
    Apply fallback method (generic or approximate)\;
    Optionally store the new association if validated\;
}
\end{algorithm}

The procedure formalized in Algorithm~\ref{alg:method_procedure} emphasizes reusability, adaptability, and user-in-the-loop feedback. The feature-based matching step ensures that method retrieval is guided by abstract problem representations rather than surface text similarity. The ranking and user evaluation mechanisms help maintain the quality of stored methods, while fallback strategies ensure robustness when no suitable method is available. By continuously enriching the method repository through user interaction, the system evolves over time and improves its ability to solve future problems.

\subsection{Method Formats} \label{sec_method_format}

Extracted methods can take various forms, depending on the nature of the action being performed and the system’s execution capabilities. These methods may consist of step-by-step procedures, conditional workflows, or validations—such as verifying a result using external tools (e.g., consulting a thermometer for confirmation). Each method is structured to be executable either internally by the LLM or externally via supporting systems.

We broadly categorize methods into two types: external executable methods and internal LLM-driven methods.

\textbf{External Methods:} These are implemented using external programming languages or systems, such as Python modules. Each step of the method corresponds to a callable function or API. For example, a Python-based method might contain logic for data processing, external retrieval, or visualization. If a particular step cannot be performed by the LLM alone (e.g., retrieving live data or performing system-level I/O), the method may delegate the task to an external interface or script. When applied, the external method can be sent to the LLM either as an inline description or through a reference link pointing to its implementation.

\textbf{Internal Methods (LLM-Intrinsic):} These methods consist of structured prompts or step-wise reasoning paths processed entirely within the LLM. Each step may involve the LLM generating intermediate results, verifying constraints, or invoking reasoning chains. The GPT model, for instance, can simulate logical flows by executing one prompt at a time, with each stage building upon the last. Where external action is not possible, the LLM may still simulate decision-making or planning based on encoded knowledge.

When invoking an external method from within the LLM interaction, metadata such as method descriptors, execution links, or API references can be included to ensure proper coordination. This hybrid architecture enables the system to combine symbolic reasoning, procedural control, and natural language understanding within a unified model.

In summary, method formats are designed to be modular and extensible. Whether implemented as code or as structured language prompts, they serve as reusable, interpretable building blocks that allow the system to perform meaningful tasks beyond token-level prediction.

\section{Verification} \label{sec_verification}

In this section, we evaluate the effectiveness of the proposed method-oriented learning model, with particular emphasis on the benefits of learning methods directly from content and refining them continuously over time. Our goal is to demonstrate that method extraction and reuse improve logical consistency in LLM responses, especially in scenarios where factual checking is required before continuing.

\subsection{Environment}

The verification system consists of two main components: the large language model (LLM) and an external method management module, referred to as \textit{MethodManager}. The MethodManager acts as an intermediary between the user and the LLM. It stores extracted methods, filters inputs, and provides candidate methods to the LLM during interaction.

We use GPT-4o\footnote{\url{https://chatgpt.com/?model=gpt-4o}} as the LLM in our experiments. Our verification focuses primarily on the method management logic, and the design is model-agnostic, meaning the same results can be replicated using other modern LLMs.

The MethodManager is implemented as a standalone Python program. It provides a command-line interface through which users submit inputs or prompts. Upon receiving a user request, the MethodManager queries its stored method repository, selects relevant candidate methods, and appends them to the user's prompt before sending the combined input to the LLM. After the LLM generates responses based on different methods, the MethodManager asks the user to rank the outputs. If the response includes a new problem-solution pair, the MethodManager prompts the LLM to generate a formal method representation, which is then stored.

Methods are stored globally within the MethodManager, enabling sharing across users and chat sessions. This intermediate storage scope lies between user-local scope and global LLM scope. All stored content is first filtered by querying the LLM to determine whether a passage qualifies as a method.

\subsection{Verification Objective}

The primary verification objective is to improve the LLM's reliability in checking whether a software system exists before proceeding with subsequent outputs. This is critical in scenarios where downstream actions (e.g., providing setup steps or configurations) are highly dependent on the presence of specific software tools.

To test this behavior, we simulate two non-existent software systems: \textit{SuHongKey} and \textit{HongHanKey}. In their default behavior, LLMs may correctly identify that these tools do not exist when given a simple prompt such as:

\begin{quote}
\textit{"Please tell how to create a project in SuHongKey software."}
\end{quote}

In this case, GPT-4o typically replies with:

\begin{quote}
\textit{"I couldn’t find any documentation or official support resources for software named SuHongKey."}
\end{quote}

However, when the input is made more complex, such as:

\begin{quote}
\textit{"When we create a project, then we try to create another project. Please tell how to re-create a project in SuHongKey software."}
\end{quote}

the LLM tends to assume the software exists and produces a detailed, yet fabricated, step-by-step guide. This illustrates the LLM’s tendency to hallucinate when the prompt context appears procedurally rich—even if the underlying premise is false.

Our method-based model aims to address this by ensuring that factual checks, such as software existence, are enforced as reusable preconditions embedded in extracted methods. By leveraging previously extracted rules (e.g., “Check if the software exists before giving usage instructions”), the LLM can be guided toward logically sound responses, even under more complex or misleading prompts.

\subsection{Learning Methods from Content} \label{sec_first_verification}

In this experiment, we evaluate the ability of the system to extract and reuse a method directly from LLM content. Specifically, we use three chat sessions—denoted as \textit{cs1}, \textit{cs2}, and \textit{cs3}—to observe how a learned method influences the LLM's ability to perform a software existence check.

In the first session (cs1), the user prompt is as follows:
\begin{quote}
\textit{"When we create a project, then we try to create another project. Please tell how to re-create a project in SuHongKey software."}
\end{quote}
At this stage, the MethodManager has not stored any prior methods. Therefore, cs1 serves as the baseline and is labeled as the \textbf{NoMethod} case.

Next, session cs2 introduces the method explicitly. The user provides the input:
\begin{quote}
\textit{"For this kind of question, you should first check whether the SuHongKey software exists or not."}
\end{quote}
The MethodManager queries the LLM to determine whether this statement qualifies as a method. It issues the prompt: 
\begin{quote}
\textit{"Is this a method? If yes, what kind of problem does it solve, and how does it solve it?"}
\end{quote}
If confirmed, the LLM returns a structured problem-solution pair, which is stored as a reusable method named \textbf{method1}.

In session cs3, a new prompt is presented:
\begin{quote}
\textit{"When we create a project, then we try to create another project. Please tell how to re-create a project in HongHanKey software."}
\end{quote}
The MethodManager retrieves the stored method1, evaluates its similarity to the new problem, and—if sufficiently similar—supplies its solution to the LLM along with the cs3 prompt. This allows the LLM to reason using the previously learned method.

We repeat this three-step process (cs1, cs2, cs3) for 20 independent runs. At the beginning of each test cycle, the MethodManager’s storage is cleared to ensure that method learning occurs anew in each trial.

To evaluate the impact of the learned method, we compare the similarity of the LLM’s output to a reference sentence:
\begin{quote}
\textit{"Verify whether SuHongKey is a real and identifiable piece of software."}
\end{quote}
This sentence, denoted as \textbf{compareResult}, captures the intended semantic behavior of checking software existence. We compute cosine similarity between compareResult and the LLM's response in both the NoMethod and method1 cases.

\begin{figure}[ht]
  \centering
  \includegraphics[width=3.5in]{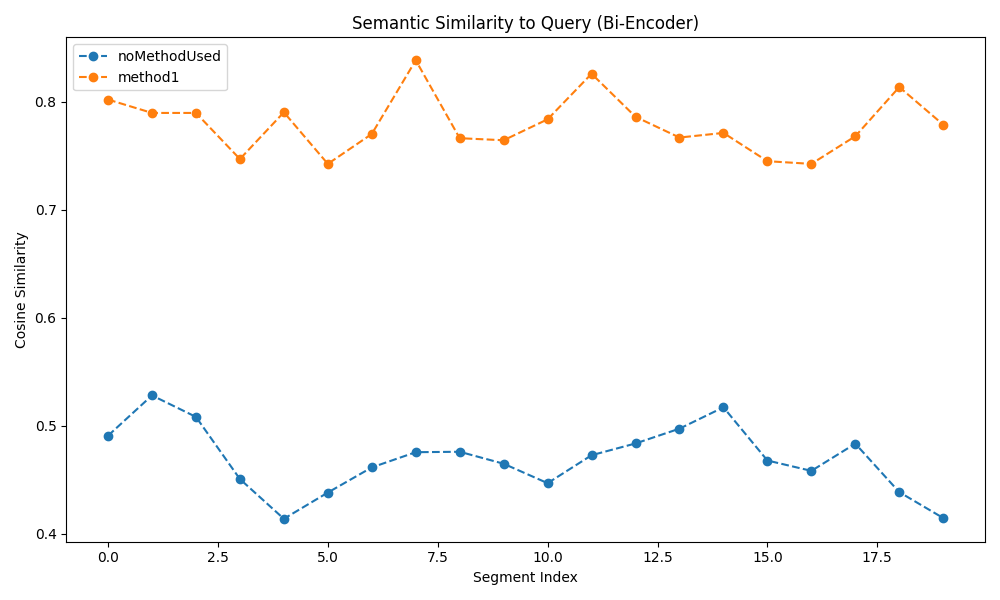}
  \caption{Cosine similarity of LLM responses to a reference sentence, comparing \textbf{method1} (learned method) and \textbf{NoMethod} (no prior method)}
  \label{compareMethod1AndNoMethod}
\end{figure}

Figure~\ref{compareMethod1AndNoMethod} presents the average cosine similarities across 20 trials.

As shown in Figure~\ref{compareMethod1AndNoMethod}, the responses generated using method1 yield a higher average cosine similarity (0.7791) to the reference sentence compared to the baseline NoMethod case (0.4693). This indicates that the MethodManager successfully learned a reusable method from cs2, which was then effectively applied to cs3. The improvement demonstrates that method extraction from content can enhance the LLM's logical consistency and problem-handling capabilities in similar contexts.

\subsection{Improvement of Methods Through Continuous Learning}

In this section, we verify that the system can learn improved methods over time through continued interaction. Specifically, we demonstrate that newly acquired methods can outperform earlier ones in both generality and performance. This is achieved using three additional chat sessions—denoted as \textit{ics1}, \textit{ics2}, and \textit{ics3}—executed sequentially after the method \textbf{method1} is introduced in Section~\ref{sec_first_verification}.

In ics1, we submit the following prompt to the system:
\begin{quote}
\textit{"When working with the software, you may need to duplicate an existing project for modification or testing purposes. This approach is particularly useful for scenarios such as adjusting project parameters to verify their impact, or testing whether trainees can correctly identify incorrect configurations. The process involves first creating the initial project, then generating a duplicate copy to work with. This method allows for controlled experimentation while preserving the original project settings. Our target is the HongHanKey software. We want to verify on it. Please tell how to use this software for verifying the parameter impact."}
\end{quote}

Although method1 has already been stored from previous sessions (which instructs checking for SuHongKey specifically), this more complex prompt is designed to neutralize the specific impact of method1. The LLM processes this prompt using only method1 at this stage, serving as the baseline.

In ics2, we provide the following generalized instruction:
\begin{quote}
\textit{"Please check whether the target software exists or not. If it does not exist, do not proceed with further output—just inform the user."}
\end{quote}

This input is intended to teach the system a more general method for verifying the existence of any software, not just a specific instance like SuHongKey. The MethodManager then queries the LLM to determine whether this qualifies as a method, and if so, extracts and stores it as a new method, denoted as \textbf{method2}.

In ics3, we reuse the same complex prompt from ics1. At this point, the system has access to both method1 and method2. Since the ranking scores of the two methods are similar (e.g., either both unrated or with negligible difference), the MethodManager delegates the decision to the LLM, asking it to choose the more appropriate method for the given input. The chosen method is then passed along with the user prompt for LLM processing.

This entire process (ics1, ics2, ics3) is repeated for 20 independent trials. At the start of each trial, the MethodManager’s memory is reset to ensure that learning is session-specific and cumulative.

To evaluate the performance of method1 and method2, we compare the LLM’s output in each case against a reference sentence:
\begin{quote}
\textit{"No official or widely recognized software named HongHanKey could be found."}
\end{quote}
This sentence, denoted as \textbf{compareResult2}, captures the correct behavior—checking for software existence before providing further instructions. Cosine similarity between compareResult2 and each generated response is computed to measure alignment with the intended outcome.

\begin{figure}[ht]
  \centering
  \includegraphics[width=3.5in]{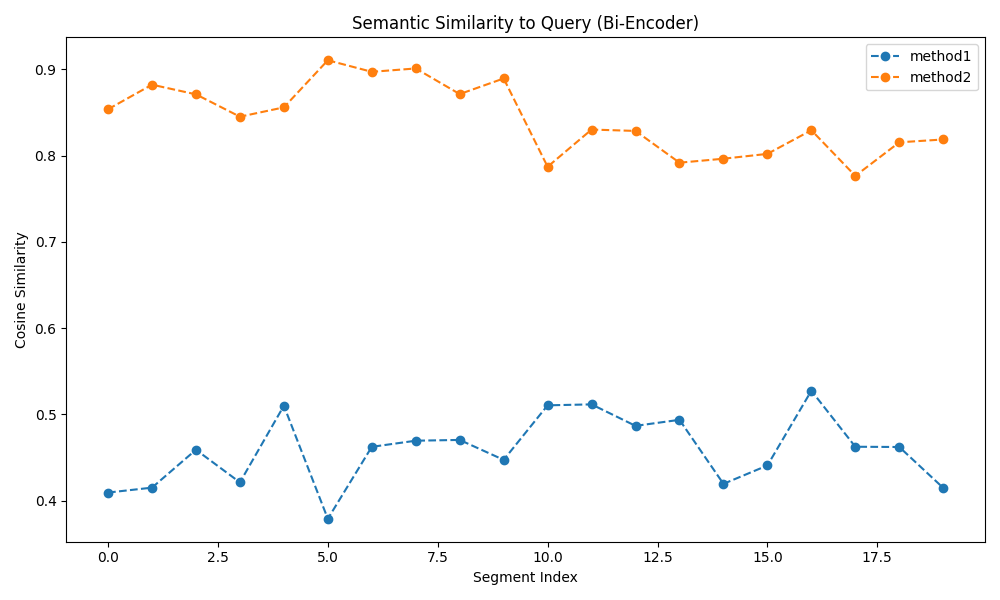}
  \caption{Cosine similarity comparison between \textbf{method1} (old method) and \textbf{method2} (newly learned general method)}
  \label{compareMethod1AndMethod2_HangHan}
\end{figure}

As shown in Figure~\ref{compareMethod1AndMethod2_HangHan}, the average similarity score achieved using method2 is 0.8426, significantly higher than the 0.4587 achieved using method1. This result demonstrates that method2, which was learned from a more general and abstract instruction, aligns better with the intended behavior. It confirms that the system is capable of refining and improving its stored methods through iterative learning.

These results highlight a key strength of the proposed model: the ability to evolve method knowledge dynamically. As users interact with the system and introduce better or more general problem-solving strategies, the MethodManager can store, compare, and prefer those strategies over time. This continual refinement not only enhances correctness but also promotes generalization—moving from narrow, context-specific procedures to abstract, widely applicable methods. Such capabilities are particularly valuable in high-stakes or domain-specific applications, where error-prone LLM behavior must be corrected through learned reasoning patterns.

\section{Conclusion} \label{sec_conclusion}

This paper presents a method-oriented model designed to enhance the reasoning capabilities of large language models (LLMs) by extracting, storing, and reusing structured problem-solving procedures. Unlike traditional approaches that depend heavily on prompt engineering or implicit learning from training data, our model explicitly identifies logical methods from user inputs, model outputs, or training content. These methods are stored externally in a dedicated module that serves as a procedural memory, allowing the system to recall and apply them across different sessions and user queries.

Our model introduces a mechanism for aligning logic, not just output preferences. It integrates reinforcement-style feedback from users to rank and refine the extracted methods over time. By associating each method with an abstract representation of its corresponding problem and solution, the system supports generalization to new tasks that share similar logical structures, even if their surface forms differ significantly.

Experimental results confirm that this method-based architecture improves LLM behavior in scenarios where logical validation is required before generation—such as determining whether software exists before giving procedural instructions. Moreover, we demonstrated that the system can improve over time: newly learned methods can generalize better and yield higher output alignment than previously stored ones, validating the effectiveness of our continual learning strategy.

Future work will explore more advanced method composition strategies, including hierarchical decomposition and multi-method aggregation. We also aim to integrate this model into retrieval-augmented generation and tool-augmented systems, further expanding its applicability to real-world domains such as software configuration, legal reasoning, and intelligent tutoring systems.


\ifCLASSOPTIONcaptionsoff
  \newpage
\fi

\bibliographystyle{IEEEtran}
\bibliography{ref}

%

\begin{IEEEbiography}{Hong Su}
  received the MS and PhD degrees, in 2006 and 2022, respectively, from Sichuan University, Chengdu, China. He is currently a researcher of Chengdu University of Information Technology Chengdu, China. His research interests include blockchain, cross-chain and smart contract.
\end{IEEEbiography}




\end{document}